\documentclass{article}

\PassOptionsToPackage{}{natbib}
\usepackage{natbib}
\usepackage[final]{nips_2018}

\usepackage[utf8]{inputenc} % allow utf-8 input
\usepackage[T1]{fontenc}    % use 8-bit T1 fonts
\usepackage{hyperref}       % hyperlinks
\usepackage{url}            % simple URL typesetting
\usepackage{booktabs}       % professional-quality tables
\usepackage{amsfonts}       % blackboard math symbols
\usepackage{nicefrac}       % compact symbols for 1/2, etc.
\usepackage{microtype}      % microtypography
\usepackage{amsmath}
\usepackage{tikz}
\usetikzlibrary{matrix,arrows}
\usepackage{xypic}

\newcommand*{\m}[1]{\mathbf{#1}}

\DeclareMathOperator*{\argmin}{arg\,min}
\usepackage[all,cmtip]{xy}

\title{Variational learning across domains with triplet information}

\author{
  Rita Kuznetsova$^{1, 2}$, Oleg Bakhteev$^{1, 2}$ and Alexandr Ogaltsov$^{2, 3}$\\
  $^{1}$Moscow Institute of Physics and Technology\\
  $^{2}$Antiplagiat Company\\
  $^{3}$National Research University Higher School of Economics \\
  \texttt{\{rita.kuznetsova, bakhteev\}@phystech.edu},  \texttt{avogaltsov@edu.hse.ru} \\
}

\begin{document}

\maketitle

\begin{abstract}
The work investigates deep generative models, which allow us to use training data from one domain to build a model for another domain. We propose the Variational Bi-domain Triplet Autoencoder (VBTA) that learns a joint distribution of objects from different domains. We extend the VBTAs objective function by the relative constraints or triplets that sampled from the shared latent space across domains. In other words, we combine the \textit{deep generative models} with a  \textit{metric learning} ideas in order to improve the final objective with the triplets information. The performance of the VBTA model is demonstrated on different tasks: image-to-image translation, bi-directional image generation and cross-lingual document classification.
\end{abstract}

\section{Introduction}
Inspired by works~\cite{karaletsos2015bayesian}, ~\cite{NIPS2014_5352}, ~\cite{suzuki2016joint}, ~\cite{vedantam2017generative} we propose Variational Bi-domain Triplet Autoencoder (VBTA) that learns a joint distribution of objects from different domains $\m{X}$ and $\m{Y}$ having a similar structure (e.g. texts, images) with the help of probabilistic triplet modeling. But, unlike these works, we suppose other form of approximate posterior distributions and sampled third triplet object across domains during training process. We suppose that on each training epoch the information from the triplets regularizes our objective. VBTA allows using distributed representations as samples from shared latent space $\m{z}$ that captures characteristics from both domains. We make assumptions about shared-latent space, in which the paired objects (images, sentences) from different domains are close to each other.
The main contributions of this paper are the following:
\begin{itemize}
\item We introduce the Variational Bi-domain Triplet Autoencoder (VBTA)~---  new extension of variational autoencoder that trains a joint distribution of objects across domains with learning triplet information. We propose negative sampling method that samples from the shared latent space purely unsupervised during training. 
\item We demonstrate the performance of the proposed model on different tasks such as bi-directional image generation, image-to-image translation, cross-lingual document classification.
\end{itemize}

\section{Assumptions}\label{assumptions}
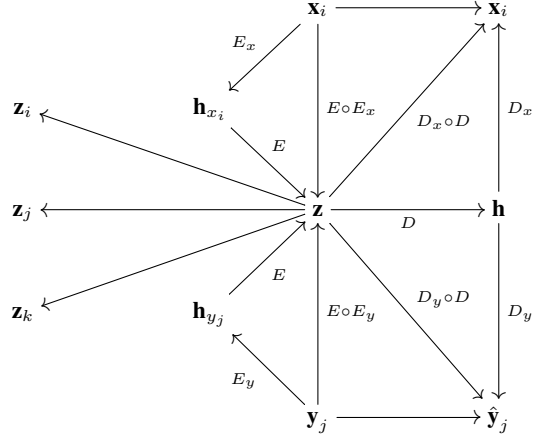
\begin{figure}[tbh!]
\small
%\begin{displaymath}
\[ \xymatrix{
    &&     & \textbf{x}_i \ar[dl]_{E_x}\ar[dd]^{E\circ E_x}\ar[rr] && \hat{\textbf{x}}_i\\
\textbf{z}_i && \textbf{h}_{x_{i}}\ar[dr]^E && \\
\textbf{z}_j &&& \textbf{z}\ar[rr]_D\ar[uurr]_{D_x \circ D}\ar[ddrr]^{D_y \circ D}\ar[lll]\ar[lllu]\ar[llld] && \textbf{h} \ar[uu]_{D_x}\ar[dd]^{D_y}\\
\textbf{z}_k && \textbf{h}_{y_{j}} \ar[ur]_E && \\
    &&     & \textbf{y}_j\ar[ul]^{E_y}\ar[rr]\ar[uu]_{E\circ E_y} && \hat{\textbf{y}}_j
     } \] 
\caption{VBTA generative process}\label{Fig1}     
%\end{displaymath}
\end{figure}

Consider dataset $(\m{X}, \m{Y}) = \{\m{x}, \m{y}\}_{n=1}^N$ consisting of $N$ \textit{i.i.d.}  objects from different domains. We assume that these objects are generated independently by the random process using the same latent variable $\m{z}$. 
We make an assumption that for each pair $(\textbf{x}, \textbf{y})$ there exists a shared latent space variable $\textbf{z}$, from which we can reconstruct both $\textbf{x}$ and $\textbf{y}$. Latent space variable $\textbf{z}$ is built from the domain space variables $\textbf{h}_x$, $\textbf{h}_y$ according to equations:
$\textbf{z} = E(\textbf{h}_{x_{i}}) = E\left(E_x(\textbf{x}_i)\right)$, $\textbf{z} = E(\textbf{h}_{y_{j}}) = E\left(E_y(\textbf{y}_j)\right)$,
where $\textbf{h}_{x_{i}}$ and $\textbf{h}_{y_{j}}$ are produced from $\textbf{x}_i$ and $\textbf{y}_j$ accordingly: $\textbf{h}_{x_{i}} = E_x(\textbf{x}_i)$, $\textbf{h}_{y_{j}} = E_y(\textbf{y}_j)$.
We define a shared intermediate variable $\textbf{h}$, which is used to obtain corresponding domain variables $\hat{\textbf{x}}_i$, $\hat{\textbf{y}}_j$ from $\textbf{y}_j$, $\textbf{x}_i$ through $\textbf{z}$: $\textbf{h} = D(\textbf{z}) = D\left(E(E_x(\textbf{x}_i))\right) = D\left(E(E_y(\textbf{y}_j))\right)$.
\[\hat{\textbf{y}}_j = D_y(\textbf{z}) = D_y\left(D(E(E_x(\textbf{x}_i)))\right) = f(\textbf{x}_i) \approx \textbf{y}_j,\hat{\textbf{x}}_i = D_x(\textbf{z}) = D_x\left(D(E(E_y(\textbf{x}_i)))\right) = g(\textbf{y}_j) \approx \textbf{x}_i.\]
The necessary condition for $f$ and $g$ to exist is the cycle-consistency constraint. That is, the proposed assumptions requires the cycle-consistency assumption.
The following diagram on Figure ~\ref{Fig1} presents VBTA generative process. Objects $\textbf{z}_i$, $\textbf{z}_i$ and $\textbf{z}_k$ form triplet.

\section{Variational Bi-domain Triplet Autoencoder}
The marginal likelihood defined by this model is:
\begin{equation}
p(\m{x}, \m{y}, \m{t}) = \int_{\m{z}} p_{\theta_{\m{x}}} (\m{x}|\m{z}) p_{\theta_{\m{y}}}(\m{y}|\m{z}) p(\m{t}_{i, j, k} | \m{z}_i,\m{z}_j,\m{z}_k) p(\m{z})  d\m{z}
\end{equation}
We can assume the following generative process:
\begin{itemize}
\item generate $\m{z}$ from prior distribution $p(\m{z}) = \mathcal{N}(\m{0}, \m{I})$,
\vspace{2pt}
\item value $\m{x}$ is generated from some conditional distribution $p_{\theta_{\m{x}}} (\m{x}|\m{z})$,
\vspace{2pt}
\item value $\m{y}$ is generated from some conditional distribution $p_{\theta_{\m{y}}} (\m{y}|\m{z})$.
\end{itemize}
%Problem of maximization the variational lower bound on the logarithm of the marginal likelihood:
%\[
%\mathcal{L}_{VBTA}(\m{x}, \m{y}, \m{t}, \phi, \theta) \rightarrow \max_{\phi, \theta}
%\]
The lower bound of the log-likelihood:
\begin{multline}
\mathcal{L}_{VBTA} = \mathbb{E}_{q_{{\phi}_x}(\m{z}_x|\m{x})} \log \frac{p_{{\theta}_x}(\m{x}, \m{y}, \m{t}, \m{z}_x)}{q_{{\phi}_x}(\m{z}_x|\m{x})} + \mathbb{E}_{q_{{\phi}_y}(\m{z}_y|\m{y})} \log \frac{p_{{\theta}_y}(\m{x}, \m{y}, \m{t}, \m{z}_y)}{q_{{\phi}_y}(\m{z}_y|\m{y})} = \\
= \underbrace{-\Bigl[KL \bigl(q_{\phi_{\m{x}} (\m{z}_x|\m{x})} (\m{z}_x|\m{x}) \parallel p_{\theta_{\m{x}}}(\m{z}_x) \bigr) + KL\bigl(q_{\phi_{\m{y}} (\m{z}_y|\m{y})} (\m{z}_y|\m{y})\parallel p_{\theta_{\m{y}}}(\m{z}_y)\bigr)\Bigr]}_{\text{Penalty}} + \\
+ \underbrace{\Bigl[\mathbb{E}_{q_{\phi_{\m{x}}} (\m{z}_x|\m{x})}  \bigl[\log  p_{\theta_{\m{x}}}(\m{x}|\m{z}_x)\bigr] 
+ \mathbb{E}_{q_{\phi_{\m{y}}} (\m{z}_y|\m{y})} \bigl[\log p_{\theta_{\m{y}}}(\m{y}|\m{z}_y)\bigr]\Bigr]}_{\text{Reconstruction}} + \\
+ \underbrace{\Bigl[\mathbb{E}_{q_{\phi_{\m{x}}} (\m{z}_x|\m{x})}  \bigl[\log  p_{\theta_{\m{x}}}(\m{y}|\m{z}_x)\bigr] 
+ \mathbb{E}_{q_{\phi_{\m{y}}} (\m{z}_y|\m{y})} \bigl[\log p_{\theta_{\m{y}}}(\m{x}|\m{z}_y)\bigr]\Bigr]}_{\text{Cycle-consistency}} + \\
+ \underbrace{\mathbb{E}_{q_{\phi_{\m{x}}} (\m{z}_x|\m{x})} \bigl[\log p(\m{t}|\m{z}_x)\bigr] + \mathbb{E}_{q_{\phi_{\m{y}}} (\m{z}_y|\m{x})} \bigl[\log p(\m{t}|\m{z}_y)\bigr]}_{\text{Triplet likelihood}}
\end{multline}

Both $q_{\phi_{\m{x}} (\m{z}_x|\m{x})}(\m{z}_x|\m{x})$ and $q_{\phi_{\m{y}} (\m{z}_y|\m{y})}(\m{z}_y|\m{y})$ are encoders, $ p_{\theta_{\m{x}}}(\m{x}|\m{z}_x)$ and $ p_{\theta_{\m{y}}}(\m{y}|\m{z}_y)$ are decoders, modeled by the deep neural networks. Similar to~\cite{DBLP:journals/corr/LiuBK17} our decoders and encoders use the common functions $E$ and $D$, see ~\eqref{assumptions}. We apply the Stochastic Gradient Variational Bayes (SGVB) and optimize the variational parameters $\theta_{\m{x}}$, $\theta_{\m{y}}$, $\phi_{\m{x}}$ and $\phi_{\m{y}}$.

\section{Learning Triplets}\label{triplets}
Based on the metric learning approach and similar to ~\cite{karaletsos2015bayesian} we extend our model by relative constraints or triplets: $\mathcal{T} = \{(\m{z}_i,\m{z}_j,\m{z}_k) \text	{ : $d$($\m{z}_i$, $\m{z}_j$) $<$ $d$($\m{z}_i$, $\m{z}_k$)} \},$ but in our case we sampled triplets across domains $\m{X}$ and $\m{Y}$.
We define the conditional triplet likelihood in the following form:
\begin{equation}
p(t_{i, j, k} = True | i, j, k) = \int_{\m{z}} p(\m{t}_{i, j, k} | \m{z}_i,\m{z}_j,\m{z}_k)  p(\m{z}_i)  p(\m{z}_j)   p(\m{z}_k) d\m{z}_i  d\m{z}_j d\m{z}_k,
\end{equation}

that was modelled by Bernoulli distribution over the states \textit{True} and \textit{False} parametrized with the use of softmax-function
\begin{equation}
p(t_{i, j, k} | i, j, k) = \frac{e^{-d(\m{z}_i,\m{z}_j)}}{e^{-d(\m{z}_i,\m{z}_j)} + e^{-d(\m{z}_i,\m{z}_k)}}
\end{equation}
Triplets~--- three objects from shared latent space $\m{z}$. $\m{z}_i$, $\m{z}_j$~--- shared latent representation of objects from $\m{X}$ and $\m{Y}$ domains.
The third object $\m{z}_k$ is sampled from domain $\m{y}$ with the minimal distance function to the corresponding objects from domain $\m{x}$ (and vice versa): 
\begin{equation}
\m{z}_{k}=\argmin_{\m{z}_{i^{'}} \in \mathcal{S}_b \backslash (\m{z}_i, \m{z}_j)}d(\m{z}_i, \m{z}_{i^{'}}),
\end{equation}
where $\mathcal{S}_b \in \mathcal{S}$~--- current mini-batch, $\mathbf{z}_i$ and $\mathbf{z}_j$~--- the paired objects from different domains.
As $d$ we use approximate form of JS-divergence, like~\cite{karaletsos2015bayesian}.
In other words, we want to choose an example $\mathbf{z}_{k}$ that is similar to $\mathbf{z}_{i}$ according to the current model parameters.

\section{Experiment and Results}
We presented the results on an image-to-image translation task: MNIST~\citet{lecun1998gradient} and CelebA~\citet{liu2015deep}. We presented results on cross-lingual text classification task on RCV1/RCV2 corpora \citet{Lewis:2004:RNB:1005332.1005345}. See architecture details in ~\eqref{pic_app}.

\subsection{Image-to-image translation for MNIST dataset}
We evaluated our approach on MNIST-transpose, where the two image domains $\m{x}$ and $\m{y}$ are the MNIST images and their corresponding transposed ones. Similar to~\cite{DBLP:journals/corr/abs-1709-06548} we used the classifier that trained on MNIST images as a ground-truth evaluator. For all the transposed images we encoded them via the model encoder $E \circ E_{\m{y}}$ and decoded via decoder $D_{\m{x}}\circ D$. Then we sent classified it. The results of the classification are shown in Table~\ref{table:mnist_acc}, where $n$ is the number of objects used for triplets sampling and cycle-consistency.

\begin{table}[h]
\small
  \caption{Classification accuracy (\%) on the MNIST-transpose
dataset. The DiscoGAN, Triple GAN and $\Delta$-GAN results are taken from ~\citet{DBLP:journals/corr/abs-1709-06548}}
  \label{table:mnist_acc}
  \centering
  \begin{tabular}{llllll}
    \bf Model & \bf $n=0$  & \bf $n=10$ & \bf $n=100$ & \bf $n=1000$ &  All\\\hline

    DiscoGAN & - & - & - & - & $15.00 \pm 0.20$ \\ \hline
    Triple GAN & - & - &$63.79 \pm 0.85$ &  $84.93 \pm 1.63$ &  $86.70 \pm 1.52$ \\ \hline
    $\Delta$-GAN & -  & - & $83.20 \pm 1.88$ &  $88.98 \pm 1.50$ &  $ 93.34 \pm 1.46$ \\ \hline
    %Proposed, different decoders & $11.32 \pm 6.15$ & $16.31 \pm 7.03$  & $81.94 \pm 6.55$ & $ 87.29 \pm 0.66$  & $89.05 \pm 0.56$ \\   \hline
 %Proposed & $18.89 \pm 3.59$ & $72.7 \pm 12.77$ & $ \mathbf{87.44 \pm 10.87}$  & $87.97 \pm 0.11$ & $89.86 \pm 0.24$ \\  \hline
 VBTA & $18.89 \pm 3.59$ & $86.57 \pm 6.338$ & $ \mathbf{90.44 \pm 0.003}$  & $\mathbf{90 \pm 0.0026}$ & $\mathbf{95 \pm 0.0006}$ \\  \hline

  \end{tabular}
\end{table}

We evaluated the marginal log-likelihood of our model on binarized versions of MNIST and MNIST-transpose. The results are listed in Table~\ref{table:mnist_vae}.
\begin{table}[tbh!]
\scriptsize	
  \caption{Marginal log-likelihood for MNIST as $\text{log}p(\mathbf{x})$ and MNIST-transpose datasets as $\text{log}p(\mathbf{y}$). 
%The JVMAE results are taken from ~\protect\cite{suzuki2016joint}. For VAE results we tested standart VAE~\eqref{vae}.
}
  \label{table:mnist_vae}
  \centering
  \begin{tabular}{lll}
    \bf Model & \bf $<{log}p(\mathbf{x})$ & $<{log}p(\mathbf{y})$\\\hline
    VAE~\cite{kingma2013auto} & -81.13 &  -81.01 \\ \hline
    JVMAE~\cite{suzuki2016joint} &  -85.35 &  -85.44 \\ \hline  
    VBTA & $\mathbf{-80.92}$ & $\mathbf{-80.91}$ \\ 
  \end{tabular}
\end{table}

\subsection{Qualitative results for CelebA dataset}
\begin{figure}[h!]
  \centering

\includegraphics[width=\textwidth]{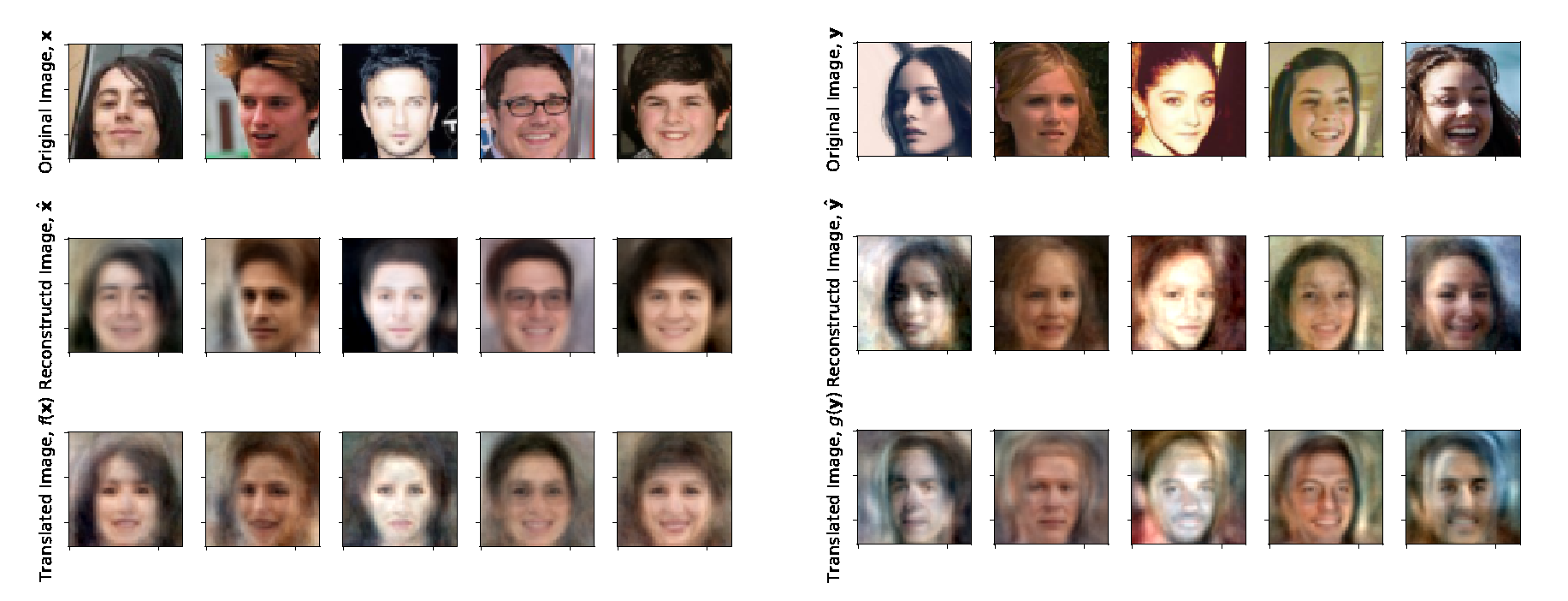}

\caption{Results of image-to-image translation for CelebA dataset.  The first row corresponds to the original images that were considered as similar because of high amount of matching attrbutes. The second row shows the reconstruction of the images. The third row illustrates the image translation from domain $\mathbf{X}$ into domain $\mathbf{Y}$ and from  $\mathbf{Y}$ into $\mathbf{X}$.}
\label{fig:faces_train}
\end{figure}

In this section we considered this dataset as a union of two domains: faces of men $\m{X}$ and faces of women $\m{Y}$.
Figure~\ref{fig:faces_train} shows the face images from datasets and their translation into different domains.

See example of bi-directional image generation in~\eqref{pair_pics}.

\subsection{Cross Lingual Document Classification}
Given a classifier trained on documents in language \textit{A} ($\mathbf{X}$ domain), one should use that classifier to predict labels of documents in language \textit{B} ($\mathbf{Y}$ domain). To handle this task we need to construct meaningful bilingual text representations. For train and test we used RCV1/RCV2 corpora, where documents are assigned to one of four predefined topics. In contrast to previous work, we \textit{do not} use parallel data at all. We artificially paired documents according to their topics.
For classification experiment, 10000 documents in English was used to train classifier and test it on 5000 documents in German and vice versa. 
Classification results are in Table~\ref{text_class_res}. See another details in~\eqref{doc_app}.

\begin{table}[h]
	\caption{Text classification accuracy}
	\label{text_class_res}
	\centering
	\begin{tabular}{p{4cm}cc}
		\toprule
		Model     & $en \rightarrow de$	& $de \rightarrow en$ \\
		\midrule
		Majority Baseline & 46.8  & 46.8     \\
		MT Baseline     & 68.1 & 67.4      \\
		\midrule
		\citet{Klementiev2012InducingCD}	& 77.6	& 71.1 \\
		\citet{Gouws:2015:BFB:3045118.3045199}	& 86.5	& 75.0 \\
		\citet{Chandar:2014:AAL:2969033.2969034}     & 91.8       & 74.2  \\
		\citet{Wei:2017:VAA:3171837.3171868}     & 92.7       & 80.4  \\
		\citet{Su:2018:NGA:3163937.3164002}	& 91.3	& 77.8 \\
		\midrule
		VBTA	& \textbf{94.3}	& \textbf{82.8} \\
		\bottomrule
	\end{tabular}
\end{table}

\section{Conclusion}
In this paper we proposed the Variational Bi-domain Triplet Autoencoder (VBTA) that learns a joint distribution of objects from different domains with the help of the learning triplets that sampled from the shared latent space across domains. We demonstrated the performance of the VBTA model on different tasks: image-to-image translation, bi-directional image generation and cross-lingual document classification. 

\textbf{Acknowledgments.} This work was supported by RFBR project No.18-07-01441

\medskip

\small

\bibliography{references}

\newpage

\appendix
\section*{Appendix A.}\label{app}

\subsection*{Image-to-image translation experiment details}\label{pic_app}

\subsubsection*{Datasets}
We used MNIST dataset for toy problem of image-to-image translation. Similar to~\cite{DBLP:journals/corr/abs-1709-06548} we considered a transposition of this dataset as a second domain $\m{y}$. 
We used 50,000 as training set and the remaining 10,000 as a test set.

CelebA consists of 202,599 face images with 40 binary attributes. 
In this work we considered this dataset as a union of two domains: faces of men $\m{x}$ and faces of women $\m{y}$.
Similar to~\cite{suzuki2016joint} we cropped and normalized the images and resized them to 64x64. Since we did not have any paired men and women in CelebA dataset, we considered that the object $\m{y}$ (women) is similar to object $\m{x}$ (men) if they had the largest matching of their attributes.

\subsubsection*{Model Architecture}\label{arch}
For the MNIST dataset we used one-layer network of 512 hidden units with ReLU for decoder $D$ and encoders $E_x, E_y$. 
For the modeling shared encoder $E$ and decoder $D_x, D_y$ we used the linear mappings. The shared latent space dimension was set to 64.

For the classification evaluation we set $p_{\theta_{\m{x}}} (\m{x}|\m{z})$ and $p_{\theta_{\m{y}}}(\m{y}|\m{z})$ to be Gaussian distribution.
For the comparison to JMVAE~\cite{suzuki2016joint} model we set $p_{\theta_{\m{x}}} (\m{x}|\m{z})$ and $p_{\theta_{\m{y}}}(\m{y}|\m{z})$ to be Bernoulli. We set model of JMVAE to the same configuration.

For CelebA we used encoders  $E_x, E_y$ with two convolution layers and a flattened layer with ReLU.
For the  shared encoder $E$ and decoder $D_x, D_y$ we used linear mapping into 64 hidden units.
For the decoder $D$ we used a network with one dense layer with 8192 units and a deconvolution layer.  
We considered $p_{\theta_{\m{x}}} (\m{x}|\m{z})$ and $p_{\theta_{\m{y}}}(\m{y}|\m{z})$ as a Gaussian distribution.

We used the Adam~\cite{kingma2014adam} optimization algorithm with a learning rate of $10^{-3}$ for the MNIST dataset and $10^{-4}$ for CelebA dataset. 
All the models were trained for 100 epochs with batch size set of 50.

\subsubsection*{Example of bi-directional image generation}\label{pair_pics}
\begin{figure}[tbh!]
  \centering
\includegraphics[width=0.5\textwidth]{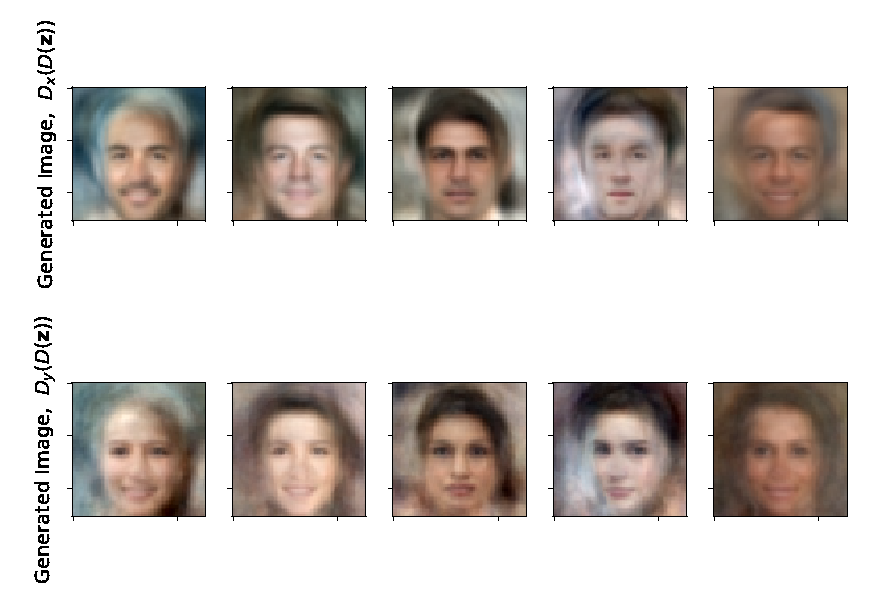}

\caption{Results of image generation from the common shared space. Each column corresponds to the faces generated from one sample of $\mathbf{z}$. The latent variable $\mathbf{z}$ was sampled from Gaussian distribution: $\mathbf{z} \sim \mathcal{N}(\mathbf{0}, \mathbf{I}).$}
\label{fig:faces_noise}
\end{figure}

Figure~\ref{fig:faces_noise} shows faces generated from Gaussian distribution. We found that our algorithm works rather well and can reproduce similar faces for both domains from one sample in latent space.

\subsection*{Cross Lingual Document Classification experiment details}\label{doc_app}
We use experimental setup similar to introduced in \citet{Klementiev2012InducingCD}. 
Previous work \citet{Chandar:2014:AAL:2969033.2969034}, \citet{Wei:2017:VAA:3171837.3171868} and \citet{Gouws:2015:BFB:3045118.3045199} used Europarl v7 parallel corpus \citet{koehn2005epc} to pretrain embeddings and then utilize it to classify subset of RCV1/RCV2 corpora \citet{Lewis:2004:RNB:1005332.1005345}. 
In this corpora documents are assigned to one of four predefined topics: CCAT (Corporate/Industrial), ECAT (Economics), GCAT (Government/Social), MCAT (Markets). In contrast to previous work, we \textit{do not} use parallel data at all. We select 15000 documents from both English and German for classification experiments.  Algorithm was trained for approximately 300K iterations with batch size equals to 50. We use Moses \citet{Koehn:2007:MOS:1557769.1557821} preprocessing tools to lowercase and tokenize texts. 
Bag-of-words was used as an initial document representation. 
We keep $30000$ top-frequency words for each language as a vocabulary.\\
We train logistic regression using low-dimensional representation obtained by our algorithm as features.

\end{document}